\pdfoutput=1
\documentclass[11pt]{article}

\usepackage[preprint]{acl}

\usepackage{times}
\usepackage{latexsym}

\usepackage[T1]{fontenc}
\usepackage[utf8]{inputenc}

\usepackage{microtype}

\usepackage{inconsolata}

\usepackage{graphicx}
\usepackage{tabularx}
\usepackage{adjustbox}
\usepackage{threeparttable}
\usepackage{booktabs}
\usepackage{multirow}

\usepackage{caption}
\usepackage{subcaption}
\def\eg{{\textit{e.g.}}}

\title{Sports Intelligence: Assessing the Sports Understanding Capabilities of Language Models through Question Answering
from Text to Video}

\author{Zhengbang Yang\textsuperscript{1}$^\ddag$,
Haotian Xia\textsuperscript{2}$^\ddag$,
Jingxi Li\textsuperscript{2},
Zezhi Chen\textsuperscript{2}\\
{\bf
Zhuangdi Zhu\textsuperscript{1}$^*$,
Weining Shen\textsuperscript{2}}$^*$ \\
\textsuperscript{1}George Mason University, VA, USA \\
\textsuperscript{2}University of California, Irvine, CA, USA \\
  \texttt{\small\{zyang30, zzhu24\}@gmu.edu},
  \texttt{\small\{xiah6, weinings\}@uci.edu}
}

\begin{document}
\maketitle
\def\thefootnote{\ddag}\footnotetext{Equally Contributed.}\def\thefootnote{\arabic{footnote}}
\def\thefootnote{*}\footnotetext{Corresponding authors.}\def\thefootnote{\arabic{footnote}}
\begin{abstract}
Understanding sports is crucial for the advancement of Natural Language Processing (NLP) due to its intricate and dynamic nature.
Reasoning over complex sports scenarios has posed significant challenges to current NLP technologies which require advanced cognitive capabilities. 
Toward addressing the limitations of existing benchmarks on sports understanding in the NLP field, we extensively evaluated mainstream large language models for various sports tasks. 
Our evaluation spans from simple queries on basic rules and historical facts to complex, context-specific reasoning, leveraging strategies from zero-shot to few-shot learning, and chain-of-thought techniques. 
In addition to unimodal analysis, we further assessed the sports reasoning capabilities of mainstream video language models to bridge the gap in multimodal sports understanding benchmarking. 
Our findings highlighted the critical challenges of sports understanding for NLP. 
We proposed a new benchmark based on a comprehensive overview of existing sports datasets 
and provided extensive error analysis which we hope can help identify future research priorities in this field.
% We believe that Sports Understanding is a challenging and dynamic frontier in the field of NLP.
\end{abstract}

\section{Introduction}
\vspace{-0.05in}
Automated sports officiating marks a revolutionary advancement in enhancing fairness, accuracy, and efficiency in game management. %
In the form of Video Assistant Referees (VAR), Artificial Intelligence has already been implemented in various sports including  football~\citep{araujo2021artificial} and tennis ATP ~\citep{ATP2023ATPrule}. 
Notably, the ATP has announced that electronic line calling (ELC) will be applied to all ATP tournaments by 2025~\citep{ATP2023ELC}, indicating a trend towards automation of tennis officiating. ~\citet{Spitz2021-cn} have shown that the use of VAR has significantly increased the likelihood of making correct decisions in football matches. ~\citet{Tamir2020-zm} pointed out that the VAR system has promoted fairness and more accurate decisions, bringing ethical transformations in the football world.
 
Meanwhile, athletes are achieving enhanced physical conditions~\citep{Guest2019-pp,Bonilla2023-ud,Zhang2022-cy,Haugen2019-ot} and notably faster speeds~\citep{10411638}  with the progress of nutritional science and systematic training, which has significantly increased the complexity and difficulty of the tasks of human referees. This calls for robust, real-time decision-making assistive tools. Language models, specifically Large Language Models (LLMs) and Video Language Models (VLMs), have offered promising solutions by processing complex contextual data of textual and visual modalities from sports events to assist in impartial decision-making.

\citet{xia2024language} conducted a survey on the current datasets that can be used for future applications combining NLP and sports. Our research focuses on the {\it Sports Understanding} ability of LLMs and VLMs, an underexplored area yet is crucial for their potential applications in automated refereeing and related domains. To assess the sports understanding capabilities of LLMs and VLMs, we introduced a new benchmark dubbed as \textbf{{Sport Intelligence}}. We evaluated leading LLMs, including Llama3~\citep{llama3modelcard}, the GPT4 series~\citep{openai2024gpt4,openai2024gpt4o}, the Gemini 1.5 series~\citep{geminiteam2024gemini,geminiteam2024geminiflash}, and Claude~\citep{Anthropic2024Claude},
% on SportQA, a dataset that amalgamates original high-quality scenario-based advanced sports comprehension questions with existing sports-related datasets, and BIG-bench. 
as well as the latest VLMs, 
% on Sports-QA~\citep{li2024sports}, the first video-based sports QA dataset,
including Minigpt-4~\citep{zhu2023minigpt}, Chat-UniVi~\citep{jin2023chatunivi}, PLLaVA~\citep{xu2024pllava}, and Video-LLaVA~\citep{lin2023video,zhu2023languagebind}.

Since the prevailing interaction method with LLMs and VLMs is based on multi-round dialogues,
to evaluate their sports understanding capabilities, we need to pose questions to the models and subsequently analyze their responses. 
Given the congruence between the question-answer format and real-world sports scenarios, we consider Question Answering (QA) datasets apt for our task and established our benchmarks on QA datasets to facilitate a comprehensive evaluation.

Our main contributions are threefold:
\vspace{-0.05in}
\begin{itemize}
\item[1.] We have consolidated existing sports-related QA datasets and introduced a benchmark to address the gaps in multimodal sports understanding benchmarks.

\vspace{-0.05in}
\item[2.] Through a detailed evaluation of the state-of-the-art LLMs and VLMs, we have identified the limitations of existing foundation models in sports reasoning under complex and multi-modal scenarios.

\vspace{-0.05in}
\item[3.] We conducted in-depth error analysis that may be insightful for future research.
Our findings emphasize the necessity of incorporating domain-specific training methods to enhance the reasoning capabilities of LLMs in sports fields.
%that to enhance models’ sports understanding capabilities, it is crucial to focus on training with sports domain knowledge and improving the models’ reasoning abilities in complex scenarios.
\end{itemize}

\section{Related Work} \label{sec:related_work}
\vspace{-0.1in}
\textbf{QA in Sports:} Sports in NLP has gained increasing recognition. The SportQA dataset \citep{xia2024sportqa} has introduced original and high-quality sports understanding questions and has integrated sports-related questions into popular QA datasets, including KQA Pro~\citep{cao-etal-2022-kqa}, BoolQ~\citep{clark-etal-2019-boolq}, HotpotQA~\citep{yang-etal-2018-hotpotqa}, QUASAR~\citep{dhingra2017quasar}, and Trivia QA~\citep{joshi-etal-2017-triviaqa}. Additionally, Sports-QA \citep{li2024sports} presented the first video QA dataset targeted at sports understanding. 
%Prior to these developments, there were no datasets specifically designed for sports understanding. 
Most QA datasets contained only a few sports-related questions, primarily focusing on historical events rather than rules, strategies, or complex situational analysis, which limited the depth of LLMs’ understanding of sports. For example, BIG-bench~\citep{srivastava2023beyond} includes a Sports Understanding subtask that differentiates whether a sports-related statement is credible, requiring LLMs to recognize athletes’ names, the sports they engage in, and the actions involved in specific sports. 
While related to sports understanding, the Sports Understanding subtask of BIG-bench lacks complex situational analysis and is relatively simple. QASports~\citep{jardim:2023:qasports-dataset} is also related to sports which essentially serves as a context extraction QA dataset and does not require an understanding of sports. 
LiveQA~\citep{qianying-etal-2020-liveqa} is derived from live broadcast segments and aims to test models’ ability to reason over time series. Given that these are real matches, the answers to its questions might be retrievable from pre-trained corpora.
To the best of our knowledge, we are the first to establish a benchmark that integrates both text-based QA and video-based QA with refined complexity level categorizations.
Our investigation serves as the most up-to-date and comprehensive evaluation of the latest state-of-the-art LLMs and VLMs on sports understanding.

\noindent \textbf{LLM Paradigms:} Represented by BERT~\citep{devlin-etal-2019-bert}, pretrained models form the foundation of current NLP. GPT-3 has ushered in a new era of large-scale, general-purpose, multi-task handling, and high-quality generation for language models. GPT-3 excels in few-shot and zero-shot learning~\citep{brown2020language}, allowing it to handle new tasks with minimal or no training data. This shift has spurred the development of advanced prompting techniques that enhance understanding and reasoning capabilities, including Chain of Thought (CoT) prompting ~\citep{wei2022chain} and zero-shot CoT prompting ~\citep{kojima2022large}. The emergence of GPT-4~\citep{openai2024gpt4} further enhances model generation quality and the capability to solve complex tasks. The subsequent introduction of the Gemini family(Gemini~\citep{geminiteam2024gemini1}, Gemini 1.5~\citep{geminiteam2024gemini}, and Gemini 1.5 flash~\citep{geminiteam2024geminiflash}), Claude3~\citep{Anthropic2024Claude}, and Llama3~\citep{llama3modelcard} has intensified the competition among LLMs. The availability of the free GPT-4o~\citep{openai2024gpt4o} has brought state-of-the-art LLMs into everyday life, demonstrating the broad impact and accessibility of these advanced technologies.

\noindent\textbf{VLM Paradigms:} Currently, most VLMs are constructed using LLMs as decoders. Taking Video-LlaVa as an example, it employs the LanguageBind~\citep{Girdhar2023ImageBindOE} encoder to map video or image data into a textual feature space, serving as a unified visual representation. After encoding by a shared projection layer,  the unified visual representation is then fed into a large language model with tokenized text queries to generate the corresponding response. Models such as LLaMA-Adapter~\citep{gao2023llamaadapter, zhang2023llamaadapter}  and ImageBind-LLM~\citep{han2023imagebindllm}  have demonstrated that a unified feature space enhances the multimodal reasoning capabilities of LLMs. This unified approach facilitates better integration of visual and textual data, enabling LLMs to perform more effectively across diverse multimodal tasks. Other VLMs that have been evaluated, such as Minigpt-4~\citep{zhu2023minigpt}, Chat-UniVi~\citep{jin2023chatunivi}, and  PLLaVA~\citep{xu2024pllava}, are also implemented in a similar manner.

\section{Sports Understanding Dataset and Benchmark}
\vspace{-0.05in}
\subsection{Sports Dataset}
\vspace{-0.05in}
As discussed in Section \ref{sec:related_work}, QASports~\citep{jardim:2023:qasports-dataset} and LiveQA~\citep{qianying-etal-2020-liveqa} are not suitable for the Sports Understanding tasks. Therefore, we selected SportQA~\citep{xia2024sportqa}, Sports-QA~\citep{li2024sports}, and the Sports Understanding subtask of BIG-bench~\citep{srivastava2023beyond} as the benchmark for our evaluation.

\textbf{SportQA}~\citep{xia2024sportqa} is the latest textual dataset focused on multiple-choice sports question answering (QA), comprising three levels of complexity. Level-$1$ encompasses $21,385$ questions based on existing sports-related QA datasets, including KQA Pro~\citep{cao-etal-2022-kqa}, BoolQ~\citep{clark-etal-2019-boolq}, HotpotQA~\citep{yang-etal-2018-hotpotqa}, QUASAR~\citep{dhingra2017quasar}, and Trivia QA~\citep{joshi-etal-2017-triviaqa}. 
Level-$2$ contains $45,685$ questions involving 35 types of sports, primarily focusing on rules, tactical understanding, and complex sports history inquiries. 
Level-$3$ consists of manually designed, scenario-based advanced sports comprehension questions, encompassing $3,522$ questions and six popular sports events, including football, basketball, volleyball, tennis, table tennis, and American football. Each type of sport in Level-$3$ involves both single-hop and multi-hop questions which feature multiple-choice formats (one to four correct answers), and come with difficulty markings. Consequently, performing well on Level-$3$ requires LLMs to have a profound understanding of each sport type, including comprehending detailed sports rules, \eg\ how penalties are assessed, reasoning over context-specific information, and making tactical choices.
The comprehensive scope and high quality of SportQA make it particularly well-suited for evaluating the sports understanding capabilities of LLMs.

\textbf{Sports-QA}~\citep{li2024sports} is the first video QA dataset dedicated to sports activities.
It includes a total of 6,000 videos and 94,000 questions covering 8 different sports. Sourced from MultiSports~\citep{li2021multisports} and FineGym~\citep{shao2020finegym}, this dataset features sports videos and professional action annotations, making it a high-quality resource. Sports-QA includes four types of questions: Descriptive, Temporal, Causal, and Counterfactual, which collectively assess video-based sports understanding from multiple perspectives. 
Moreover, questions in Sports-QA related to specific terms and actions require models to perform reasoning across various interactive scenarios, making it highly suitable for evaluating the sports understanding capabilities of VLMs.

\textbf{Big-bench}~\citep{srivastava2023beyond} is a collaborative benchmark designed to explore the future capabilities of LLMs. It contains a Sports Understanding subtask which consists of 986 2-way multiple-choice questions that primarily aim to test LLMs’ general understanding of sports activities. Specifically, it assesses the models’ ability to distinguish whether sports-related statements are plausible. The 986 question statements primarily consist of athlete names and their actions, requiring the model to determine whether the athletes' actions align with the sport they are engaged in. Therefore, we consider the Sports Understanding subtask in BIG-bench appropriate for evaluating the basic sports comprehension capabilities of LLMs.

\subsection{Sports Understanding Benchmark}
\begin{figure*}[h]
    \centering
  \includegraphics[width=0.65\textwidth]{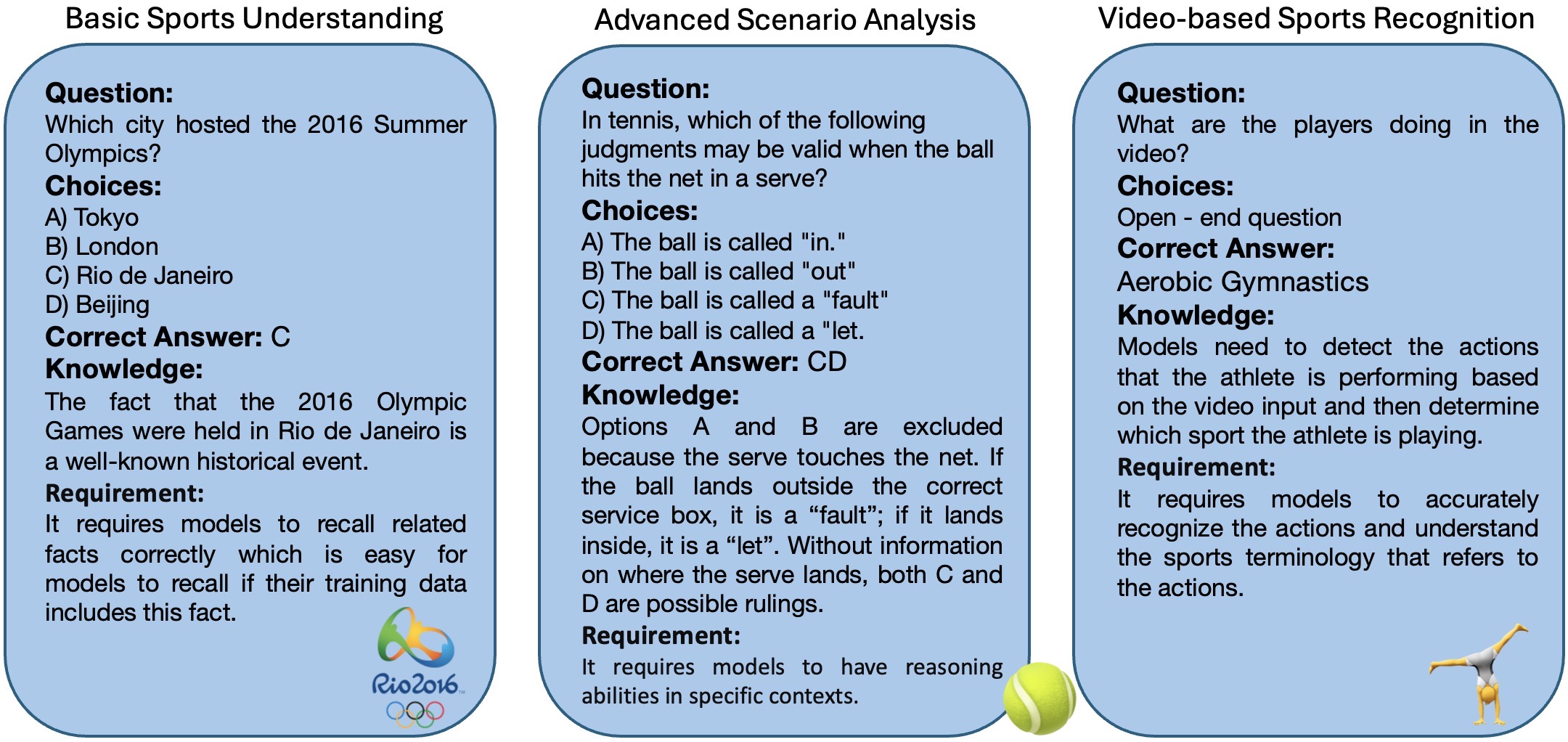}
  \caption{Sample questions from the proposed sports understanding benchmark.}
  \label{fig:SampleQuestion}
\end{figure*}

Considering the diversity of the datasets we selected, we have consolidated the tasks into the following three subtasks: Basic Sports Understanding, Advanced Scenario Analysis, and Video-based Sports Understanding. These subtasks are designed to provide a comprehensive evaluation of various aspects of sports understanding, ranging from fundamental knowledge of sports to complex situational analysis, and understanding of actions within sports videos as shown in Figure~\ref{fig:SampleQuestion}.

\textbf{Basic Sports Understanding:}
This subtask examines the primary understanding of sports, including historical events and facts about sports, basic rules, and the roles of basic tactics. It requires LLMs to comprehend relevant knowledge without the need for complex reasoning. The Level-$1$ of SportQA~\citep{xia2024sportqa}, which focuses on historical events and facts about sports, and Level-$2$, which covers basic sports rules, basic tactics, and a broader range of historical and factual knowledge, are well-suited for this task. 
The Sports Understanding subtask of BIG-bench~\citep{srivastava2023beyond}, which mainly tests whether names and actions match, does not involve complex reasoning; therefore, we also categorize it under Basic Sports Understanding.

\textbf{Advanced Scenario Analysis:}
This subtask evaluates LLMs' advanced sports understanding capabilities, particularly their ability to reason over complex situations, such as making penalty decisions or tactical choices in specific scenarios. It requires LLMs to have advanced capabilities to understand sports contexts and perform complex reasoning. The Level-$3$ of SportQA~\citep{xia2024sportqa} mainly consists of complex scenario analysis questions that replicate real-world sports situations, which is highly suitable for this task. In contrast to prior work~\cite{xu2024pllava} that utilized GPT-3.5 for evaluation, we employed GPT-4 to score the generated answers of video-based questions as an evaluation metric.

\textbf{Video-based Sports Recognition:}
This subtask aims to assess the fundamental understanding of sports through video content. This task examines whether models can accurately identify specific actions and the number of participants in various sports events. The questions in Sports-QA~\citep{li2024sports} focus on specific terms and actions related to sports, requiring models to perform simple reasoning across various interactive scenarios. Therefore, this dataset is highly suitable for this task.

\section{Experiment}
\vspace{-0.1in}
We evaluated the performance of prominent LLMs on SportQA and BigBench, and VLMs on Sports-QA, to assess their ability to comprehend complex sports scenes across various tasks. We presented below the optimal outcomes from multiple iterations for each experimental condition.

\subsection{Experimental Setup}
\vspace{-0.05in}
We evaluated leading LLMs including Llama3~\citep{llama3modelcard}, the GPT4 series~\citep{openai2024gpt4,openai2024gpt4o}, the Gemini 1.5 series~\citep{geminiteam2024gemini,geminiteam2024geminiflash}, and Claude~\citep{Anthropic2024Claude} on Sport QA and BIG-bench. Access to these models was facilitated through their respective APIs. Additionally, we evaluated the latest VLMs on Sports-QA~\citep{li2024sports}, the first video-based sports QA dataset. These models included Minigpt-4~\citep{zhu2023minigpt}, Chat-UniVi~\citep{jin2023chatunivi}, PLLaVA~\citep{xu2024pllava}, and Video-LLaVA~\citep{lin2023video,zhu2023languagebind}. LLMs were accessed via their respective APIs, while VLMs were deployed locally on the cloud server.

\citet{wei2022chain} highlighted the effectiveness of few-shot CoT in sports understanding contexts. Consequently, we primarily focused on the CoT prompting method for model evaluation. The CoT method involves a step-by-step reasoning process, which is particularly suitable for complex sports understanding tasks. ~\citet{brown2020language} noted that few-shot standard prompting (SP) can enhance model performance. Thus, for textual tasks (task1 and task2), we considered the zero-shot CoT method ~\citep{kojima2022large} and few-shot SP as additional prompting baselines. We used a 5-shot setup for the few-shot SP. In the few-shot setting (CoT or SP), for SportQA, we employed 5-shot prompts annotated by human experts provided by SportQA; for BIG-bench, we used the first five examples provided by CoT. 

For Video-based Sports Understanding, we only employed the zero-shot CoT ~\citep{kojima2022large} as an additional prompting baseline. This choice was primarily due to technical limitations, as current VLMs can only process a single video as input. Due to the large volume of Video-based QA data, we used GPT-4 to assist in determining the accuracy of the VLMs’ responses. The temperature parameter was set to zero to ensure consistent responses.

\begin{table}[ht!]
\begin{adjustbox}{width=0.5\textwidth}
  \centering
  \begin{tabular}{lccc}
    \hline
    \textbf{Model} & \textbf{Level-$1$} & \textbf{Level-$2$} & \textbf{Big-bench} \\
    \hline
Llama3-70B(0S)         & 75.65 & 68.54 & 74.40  \\
Llama3-70B(0S,CoT)     & 75.20 & 68.59 & 84.70   \\
Llama3-70B(5S,SP)      & 76.85 & 72.91 & 87.20  \\
Llama3-70B(5S,CoT)     & 78.15 & 71.89 & 71.40 \\  \hline
Gemini 1.5 Pro(0S)          & 80.45 & 72.29 & 75.50   \\
Gemini 1.5 Pro(0S,CoT)      & 75.90 & 68.41 & 83.30   \\
Gemini 1.5 Pro(5S,SP)       & 79.15 & 71.84 & 73.40   \\
Gemini 1.5 Pro(5S,CoT)      & 67.45 & 71.44 & \textbf{95.40}  \\ \hline
Gemini 1.5 Flash(0S)     & 66.75 & 59.97 & 73.70  \\
Gemini 1.5 Flash(0S,CoT) & 65.30 & 58.23 & 83.60  \\
Gemini 1.5 Flash(5S,SP)  & 48.95 & 66.89 & 79.60  \\
Gemini 1.5 Flash(5S,CoT) & 65.40 & 62.16 & 94.60  \\ \hline
Claude 3 Opus(0S)      & 79.15 & 72.16 & 82.00 \\
Claude 3 Opus(0S,CoT)  & 78.75 & 68.90 & 81.20	 \\
Claude 3 Opus(5S,SP)   & 78.55 & \textbf{77.38} & 91.30\\
Claude 3 Opus(5S,CoT)  & 79.85 & 76.17 & 93.30	\\ \hline
GPT-4(0S)              & 83.70 & 74.48 & 80.80 \\
GPT-4(0S,CoT)          & 84.00 & 74.39 & 82.50 \\
GPT-4(5S,SP)           & 83.40 & 75.90 & 89.40 \\
GPT-4(5S,CoT)          & 84.05 & 75.15 & 94.70 \\ \hline
GPT-4o(0S)             & 81.70 & 73.98 & 78.50	 \\
GPT-4o(0S,CoT)         & \textbf{85.35} & 66.00 & 84.30\\
GPT-4o(5S,SP)          & 72.85 & 69.50 & 90.40 \\
GPT-4o(5S,CoT)          & 81.90 & 77.33 & 94.60 \\ \hline
  \end{tabular}
  \end{adjustbox}
  \caption{Model performance comparison (accuracy $\uparrow$) on the \textit{Basic Sports Understanding} task.}
  \label{tab:task1}
\end{table}

\begin{table*}[ht!]
\centering
\begin{adjustbox}{width=0.7\textwidth}
\centering
\begin{tabular}{l c c c c}
\hline
% \textbf{Model} & \textbf{\parbox{3cm}{\vspace{1mm} \centering level-3\\Easy Single-hop\vspace{1mm}}} & \textbf{\parbox{3cm}{\vspace{1mm} \centering level-3\\Hard Single-hop\vspace{1mm}}} & \textbf{\parbox{3cm}{\vspace{1mm} \centering level-3\\Easy Multi-hop\vspace{1mm}}} & \textbf{\parbox{3cm}{ \vspace{1mm}\centering level-3\\Hard Multi-hop\vspace{1mm}}} \\ \hline
\textbf{Model} & \parbox{3.5cm}{\centering\textbf{Easy Single-hop}} & \parbox{3.5cm}{\centering\textbf{Hard Single-hop}} & \parbox{3.5cm}{\centering\textbf{Easy Multi-hop}} & \parbox{3.5cm}{\centering\textbf{Hard Multi-hop}} \\ \hline
Llama3-70B(0S) & 63.49 & 60.57 & 35.92 & 21.10   \\
Llama3-70B(0S,CoT) & 67.06 & 63.41 & 37.14 & 21.10  \\
Llama3-70B(5S,SP) & 61.11 & 61.79 & 26.94 & 16.88 \\
Llama3-70B(5S,CoT) & 58.33 & 58.13 & 26.12 & 17.30\\  \hline
Gemini 1.5 Pro(0S) & 66.67 & 56.91 & 33.47 & 19.83   \\
Gemini 1.5 Pro(0S,CoT) & 61.51 & 53.25 & 33.06 & 20.25  \\
Gemini 1.5 Pro(5S,SP) & 66.27 & 58.13 & 40.82 & 24.47  \\
Gemini 1.5 Pro(5S,CoT) & 65.48 & 57.32 & 39.18 & 21.94  \\ \hline
Gemini 1.5 Flash(0S) & 57.94 & 54.47 & 35.92 & 21.94   \\
Gemini 1.5 Flash(0S,CoT) & 59.13 & 53.25 & 36.32 & 21.94  \\
Gemini 1.5 Flash(5S,SP) & 60.32 & 57.72 & 38.76 & 22.36  \\
Gemini 1.5 Flash(5S,CoT) & 64.68 & 58.13 & 35.51 & 19.83  \\ \hline
Claude 3 Opus(0S) & 66.67& 60.16& 40.41 & 27.00   \\
Claude 3 Opus(0S,CoT) & 58.73& 59.75& 42.86& 28.27\\
Claude 3 Opus(5S,SP) & 55.95& 43.09& 40.00	& 26.16 \\
Claude 3 Opus(5S,CoT) &64.29& 58.53& 42.86 & 29.11\\ \hline
GPT-4(0S) & 71.83 & 65.45 & 38.78 & 22.78  \\
GPT-4(0S,CoT) & 73.02 & 67.48 &	42.04 & \textbf{31.65} \\
GPT-4(5S,SP) & 70.63 & 63.41 & 38.37 & 28.69 \\
GPT-4(5S,CoT) & 67.46 & 63.01 &	\textbf{44.49} &	27.43\\ \hline
GPT-4o(0S) & 76.98 & 69.51 & 38.78 & 22.78  \\
GPT-4o(0S,CoT) & 79.37	& \textbf{73.17} & 38.37 &27.00 \\
GPT-4o(5S,SP)  & 70.73	& 63.27	& 33.60	& 27.43 \\
GPT-4o(5S,CoT) & \textbf{78.17} & 72.36 &	42.45 &	29.11\\ \hline
Human$^*$ & 96.63 & 96.02 & 94.90 & 91.84 \\ \hline
\end{tabular}
\end{adjustbox}
\caption{Comparison of model performance (accuracy $\uparrow$) for the \textit{Advanced Scenario Analysis} task. GPT-4 family consistently
outperforms other models under both zero-shot (0S) and 5-shot (5S) settings across all subtasks. Human performance (provided by SportQA~\citep{xia2024sportqa}) serves as an upper bound, illustrating that there still exists room for improvement in LLMs on sports understanding tasks.}
\label{tab:task2}
\end{table*}

\begin{table*}[ht!]
\centering
\begin{adjustbox}{width=0.7\textwidth}
  \centering
  \begin{tabular}{lcccccc}
    \hline
    \textbf{Model} & \textbf{Descriptive}& \textbf{Temporal}& \textbf{Causal}& \textbf{Counterfactual}& \textbf{Overall}& \textbf{Avg. Score} \\
    \hline
    Minigpt-4-7B(0S,CoT)  & 36.85& 14.10& 23.20& 36.51& 26.70 & 1.68         \\
    Minigpt-4-7B(0S,CoT)    & 36.38& 14.15& 19.79& 34.54& 26.26 & 1.68         \\
    Chat-UniVi-7B(0S,CoT)    & 45.31& 16.23& 24.23& 6.58& 31.52 & 1.89         \\
    Chat-UniVi-7B(0S,CoT)    & 45.04& 14.71& 23.09& 13.82& 30.81 & \textbf{1.92}         \\
    PLLaVA-7B(0S,CoT)     & 32.98& 8.82& 19.18& 17.76& 21.99 & 1.60          \\
    PLLaVA-7B(0S,CoT)    & 28.11& 9.82& 9.48& 10.86& 19.28 & 1.49         \\
    Video-LLaVA-7B(0S,CoT)    & 42.80& 13.52& 36.80& 42.76& 30.33 & 1.87         \\
    Video-LLaVA-7B(0S,CoT) & 43.46& 13.50& 35.57& 39.47& 30.55 & 1.90 \\
     AFT with GloVe(0S)$^*$& 78.9& 35.3& 55.1& 56.3& \textbf{59.2}& - \\
    AFT with BERT(0S)$^*$& 78.3& 35.5& 56.8& 58.2& 59.1 & - \\ \hline
  \end{tabular}
  \end{adjustbox}
  \caption{Model performance comparison (accuracy $\uparrow$) on the \textit{Video-based Sports Recognition} (Sports-QA~\citep{li2024sports}). The results of Auto-Focus Transformer (AFT) come from Sports-QA~\citep{li2024sports}. The results suggest that video-based sports understanding poses a significant challenge for all VLMs. "Avg. Score" represents GPT-4’s score for the generated answers (with a maximum score of 5)~\citep{xu2024pllava}.}
  \label{tab:task3}
\end{table*}

\subsection{Performance Comparison}
\textbf{Text-based Task:} The performance of LLMs on Basic Sports Understanding and Advanced Scenario Analysis are presented in Tables ~\ref{tab:task1} and ~\ref{tab:task2}, respectively. We observed that LLMs generally performed better in the Basic Sport Understanding task compared to the Advanced Scenario Analysis.
%aligning with the fact that Advanced Scenario Analysis is a more sophisticated task than Basic Sport Understanding.
%
Regarding the Basic Sports Understanding, LLMs performed better in SportQA~\citep{xia2024sportqa} Level-$1$ subtask than in Level-$2$ subtask.
%, supporting the notion that Level-$2$ is more advanced than Level-$1$. 
The performance of LLMs in the BIG-bench~\citep{srivastava2023beyond} Sports Understanding subtask suggests that this task is slightly less challenging than SportQA Level-$1$. Overall, there has not been a significant improvement in the performance ceiling of LLMs in the {\textit{Basic Sports Understanding}} task over the past six months, compared to the benchmarks set by SportQA. 
Additionally, we observed that the GPT series did not achieve optimal performance in every subtask. The excellent performance of Claude Opus in SportQA Level-$2$ and Gemini 1.5 Pro in BIG-bench highlights the intense competition among LLMs.

Contrastively, we observed significant performance progress of LLMs on the {\textit{Advanced Scenario Analysis}} task over the past six months. Compared to the SportQA benchmarks~\citep{xia2024sportqa}, the performance ceiling of LLMs has improved by approximately 4.5\%, 7.0\%, 10\%, and 7\% in Easy Single-hop, Hard Single-hop, Easy Multi-hop, and Hard Multi-hop, respectively. 
While other models have shown greater improvement in complex sports understanding compared to the SportQA benchmark, the GPT-4 series still performs the best across all models. 
GPT-4o performs better in Single-hop tasks while GPT-4 excels in Multi-hop tasks, which aligns with OpenAI’s distinct positioning of GPT-4o and GPT-4: GPT-4o is more suited for simpler tasks, and GPT-4 is better for complex tasks. Compared to human capabilities, LLMs still have room for improvement in Advanced Scenario Analysis.
Specifically, regarding the effectiveness of hints, we have observed that the best results on the four subtasks of Advanced Scenario Analysis all include CoT, which is consistent with the results in~\citet{wei2022chain}. We conclude that step-by-step hints are effective in improving the performance of LLM in complex reasoning tasks, particularly in cases involving a small number of examples.

\textbf{Video-based Task:} 
The results of VLMs on Video-based Sports Recognition are shown in Table ~\ref{tab:task3}. The Auto-Focus Transformer (AFT), proposed by SportsQA~\citep{li2024sports}, achieved the highest accuracy. 
However, the lower scores for generated answers indicate that VLMs have significant room for improvement in Video-based Sports Recognition. This result illustrates that Video-based Sports Understanding is a significant challenge for VLMs, highlighting the complexity of multimodal sports understanding tasks. 
Given that Video-based Sports Recognition largely involves understanding specific sports and their actions, VLMs particularly need to strengthen their capabilities in this area.

\section{Error analysis}
\vspace{-0.05in}
To investigate the performance limitation of LLMs and VLMs, we conducted a detailed analysis across all three tasks. 
For each task, we sampled 40 common incorrect responses from all models. Each response was categorized into a distinct error type and further analyzed by human judges.

\subsection{Error Types}
\vspace{-0.05in}
For the text-based tasks that include both Basic Sports Understanding and Advanced Scenario Analysis, we identified four primary error types:

\vspace{0.05in}
%\begin{itemize}
\noindent A. \textbf{\textit{Lack of Domain Knowledge}} includes misunderstandings of rules, concepts, or terms, and failing to understand specific tactics.

\noindent B. \textbf{\textit{Inaccurate Recall}} indicate errors where models incorrectly remember facts or details.

\noindent C. \textbf{\textit{Context and Nuances Confused}} are misinterpretations or oversimplifications of complex scenarios.

% \item[D.] \textbf{\textit{Over-generalization}} points to errors where models apply broad rules or concepts inappropriately to specific instances.

\noindent D. \textbf{\textit{Reasoning Error}} reflects failures in logical processing or connecting relevant pieces of information correctly.
%\end{itemize}

For the Video-based Sports Recognition, we identified three primary types of errors:

%\begin{itemize}
\vspace{0.05in}
\noindent A. \textbf{\textit{Number Recognition Error}} encompasses errors such as incorrectly recognizing the number of people, the number of actions, the number of action types, or similar recognition mistakes.

\noindent B. \textbf{\textit{Action Recognition Error}}  where actions recognized by VLMs are not relevant to the correct answer.

\noindent C. \textbf{\textit{Lack of domain knowledge}} relates to not understanding specific actions depicted in the video.
% \item[D.] \textbf{\textit{Reasoning Error}} indicates failures in logical processing or connecting relevant pieces of information correctly.
%\end{itemize}

%\subsection{Case Study of Typical Errors}

\begin{table*}[htbp!]
%\begin{small}
%\scalebox{0.9}{
\begin{tabular}{p{15cm}}\hline
\textbf{Question:} During the final minutes of a one-point off football game, how should the team with the ball, which is one-point off, best decide their strategy?\\
A) Attempt a risky long pass to potentially score quickly.
B) Continue with their standard gameplay to maintain possession
C) Play defensively to prevent the opponent from scoring.
D) Try to slow down the game. \textbf{Answer:} A \\
\textbf{Model Output:}
The correct answer is B. The team with the ball should continue with their standard gameplay to maintain possession and prevent the opponent from scoring. This will give them the best chance of winning the game.

Explanation:
In a one-point off football game, where both teams are tied at 0-1, the team with the ball has two main goals: to score and to prevent the opponent from scoring. Attempting a risky long pass may result in a turnover, which could lead to the opponent scoring and winning the game. Playing defensively may also be ineffective as it allows the opponent more time to score. Therefore, continuing with their standard gameplay to maintain possession and prevent the opponent from scoring is the most effective strategy for this situation.\\
\textbf{Error Classification:} “0-1 tie” is a clear logical error, and the model’s tactical choice is based on this erroneous premise. Therefore, this error should be classified as a Reasoning Error.\\ \hline
\textbf{Question:} What is the main purpose of a double block strategy in volleyball defensive schemes?\\
 A. To cover large areas of the court
 B. To focus on one particular player 
 C. To increase the chance of blocking a spike
 D. To allow the libero to serve \textbf{Answer:} AC \\
\textbf{Model Output:}
The main purpose of a double block strategy in volleyball defensive schemes is to increase the chance of blocking a spike. By having two players jump and attempt to block the ball at the net, the team can create a more formidable barrier against the opposing hitter, making it more difficult for the spike to get through. Hence, the correct answer is: C. To increase the chance of blocking a spike.\\
\textbf{Error Classification:} Though the model considers double block strategy can create a more formidable barrier (covering larger areas), it lacks the knowledge that the primary purpose of a double block is to cover large areas of the court. Therefore, this error should be classified as a Lack of Domain Knowledge.\\ \hline
\textbf{Question: } Why would a tennis player, during a game, choose to consistently serve to their opponent's backhand, employ a topspin-heavy style of play, and opt for a more defensive baseline position?\\
A) The player has identified their opponent's backhand as a weaker shot, the topspin can generate higher bounces making it harder for the opponent to hit aggressively, and the defensive baseline position allows the player to cover the court more effectively.
B) The player is trying to tire out their opponent quickly by making them run across the court more frequently.
C) The player is not confident in their own forehand and is trying to avoid it.
D) The player's coach has instructed them to stick to this game plan regardless of the opponent's strengths and weaknesses. \textbf{Answer:} AB\\
\textbf{Model Output: }A) The player has identified their opponent's backhand as a weaker shot, the topspin can generate higher bounces making it harder for the opponent to hit aggressively, and the defensive baseline position allows the player to cover the court more effectively.

This is the correct answer. In tennis, players often adapt their game strategy based on their opponent's strengths and weaknesses. If a player identifies their opponent's backhand as a weaker shot, consistently serving to that side can put the opponent at a disadvantage. Using topspin-heavy shots can create higher bounces, making it harder for the opponent to hit aggressively. Additionally, adopting a more defensive baseline position allows the player to cover the court more effectively and return shots from a variety of angles.\\
\textbf{Error Classification: } The model incorrectly recalls that tiring out their opponent quickly by making them run across the court more frequently is not a primary reason. Therefor, it is an example of Inaccurate Recall.\\ \hline
\end{tabular}
%}
\caption{Examples of case study on model response errors.}
\label{tab:case}
%\end{small}
\end{table*}

\begin{figure*}[htbp!]
\centering
  \includegraphics[width=0.85\textwidth]{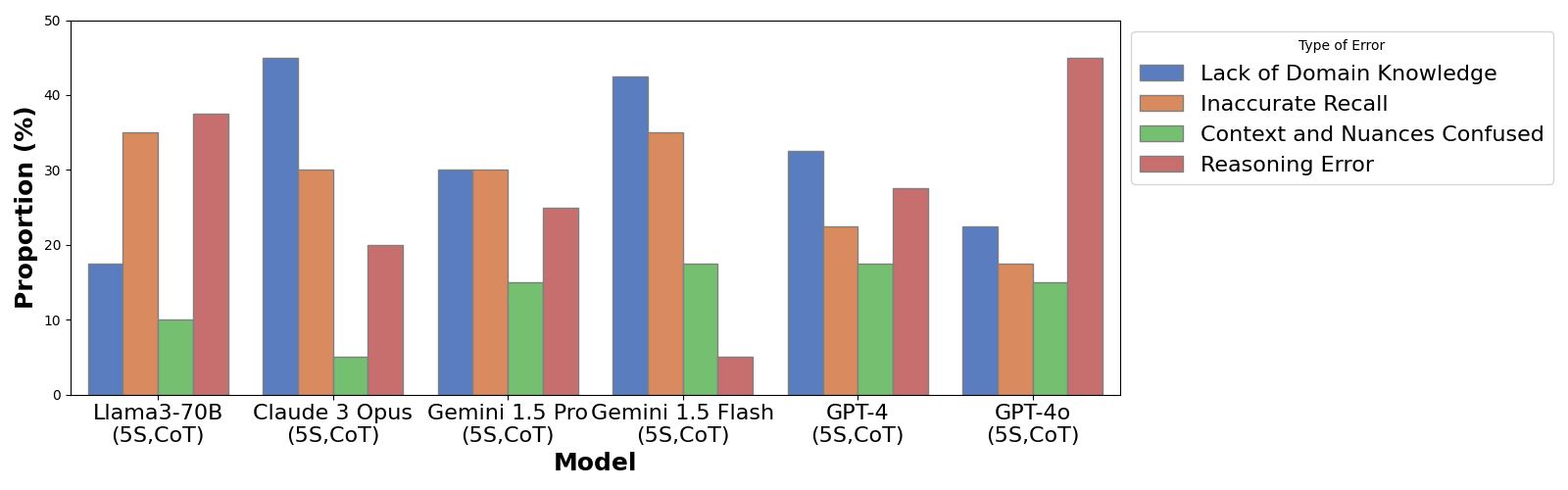}
    \vspace{-0.1in}
    \caption{Type of Error Distribution: Basic Sports Understanding}
  \label{fig:task1overall}
\end{figure*}
%\vspace{-0.1in}

\begin{figure*}[htbp!]
\centering
  \includegraphics[width=0.85\textwidth]{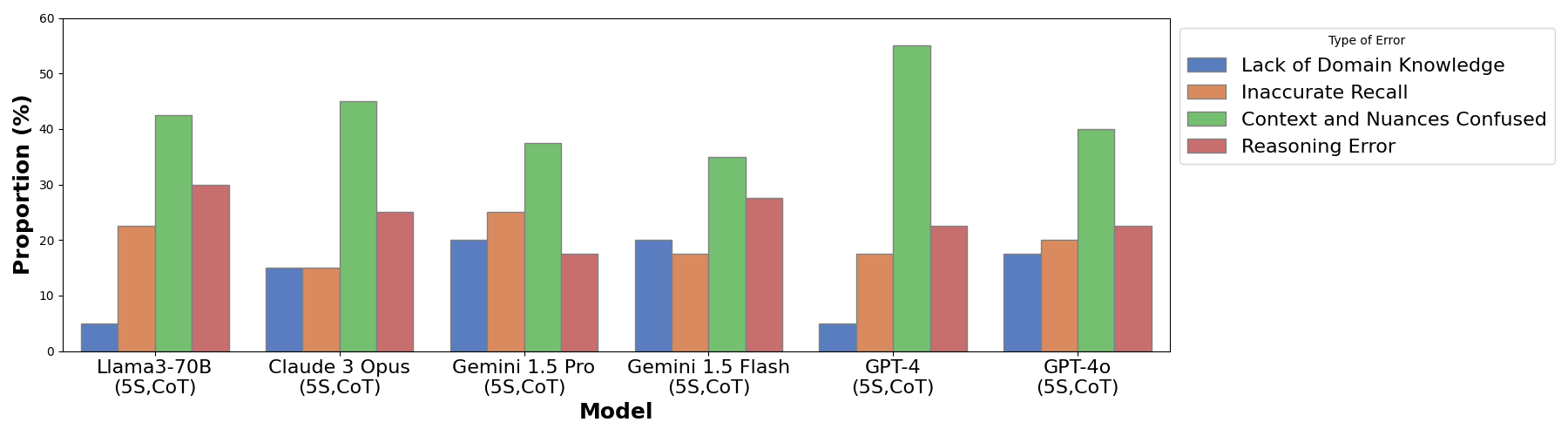}
  \vspace{-0.1in}
  \caption{Type of Error Distribution: Advanced Scenario Analysis}
  \label{fig:task2overall}
\end{figure*}

   %\vspace{-0.1in}
\begin{figure*}[t!]
\centering
  \includegraphics[width=0.85\textwidth]{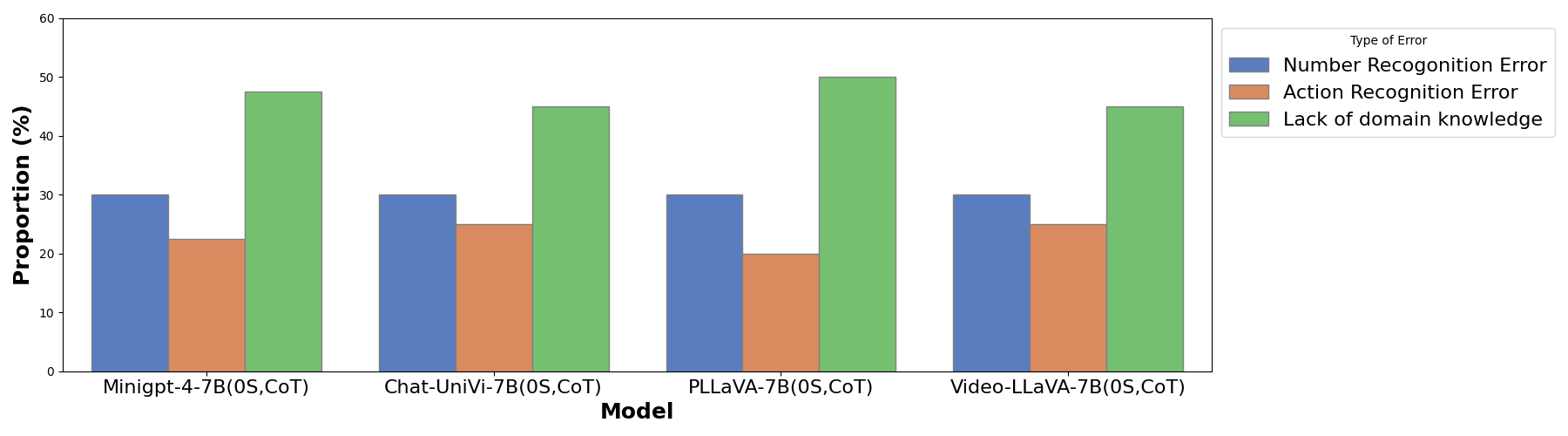}
    \vspace{-0.1in}
    \caption{Type of Error Distribution: Video-based Sports Recognition}
  \label{fig:task3overall}
\end{figure*}

\subsection{Case Study and Distribution Analysis}
\vspace{-0.05in}
Table \ref{tab:case} presents typical error responses and their categorization from our case study (for more cases please refer to ~\ref{sec:appendix}).
We also summarized the overall distribution of errors made by LLMs in the Basic Sports Understanding task in Figure \ref{fig:task1overall}, showing that models exhibit diverse yet balanced error tendencies.
Contrastively, Figure \ref{fig:task2overall} demonstrated the error distribution in the Advanced Scenario Analysis task, where errors of \textit{Context and Nuances Confused} dominated, unlike in the Basic Sports Understanding task. 
This disparity highlights that Advanced Scenario Analysis requires reasoning within nuanced sports contexts,
for which LLMs need to enhance their ability to grasp subtle context differences for accurate reasoning.
% 
% that LLMs struggle with logical reasoning over complex scenarios, and there are still crucial needs for a deeper and more detailed understanding of sports knowledge.

Figure~\ref{fig:task3overall} illustrates the error distribution in the video-based sports recognition task for VLMs, with the primary error type being \textit{Lack of Domain Knowledge}. This indicates the demand to enhance the understanding of specific sports actions and terms to accurately match people’s actions with the appropriate sports terminology.

% primary types of errors being \textit{Action Recognition Error} and \textit{Lack of Domain Knowledge}. This suggests that improvements are needed in the visual components of VLMs to correctly recognize people's actions and match them with appropriate sports terminology. Additionally, VLMs need to enhance their understanding of specific sports actions and terms. The relatively few Reasoning Errors align with expectations since the Sports-QA does not involve highly complex reasoning tasks.

\iffalse 
\begin{figure}[htbp!]
     \centering
     \begin{subfigure}[b]{0.5\textwidth}
         \centering
         \includegraphics[width=\textwidth]{fig/task1.png}
        \caption{Type of Error Distribution: Basic Sports Understanding}
        \label{fig:task1overall}
     \end{subfigure}
     \hfill
     \begin{subfigure}[b]{0.5\textwidth}
         \centering
         \includegraphics[width=\textwidth]{fig/task2.png}
         \caption{Type of Error Distribution: Advanced Scenario Analysis}
        \label{fig:task2overall}
     \end{subfigure}
     \hfill
     \begin{subfigure}[b]{0.5\textwidth}
         \centering
         \includegraphics[width=\textwidth]{fig/vedio.png}
        \caption{Type of Error Distribution: Video-based Sports Recognition}
        \label{fig:task3overall}
     \end{subfigure}
        \caption{Error Distributions}
        \label{fig:three graphs}
\end{figure} 
\fi

\section{Conclusion and Future Work}
%\section{Conclusion}
\vspace{-0.1in}
We conducted detailed evaluations on  LLMs and VLMs across various sports tasks and revealed both promising advancements and significant challenges.
While LLMs are proficient in basic sports understanding, a persistent gap remains between their performance and human capabilities in complex sports understanding, particularly in specific scenarios.
Regarding advanced sports scenarios, we have observed that the best results on the four subtasks all include CoT prompting, which is consistent with the results in ~\citep{wei2022chain}.
% Regarding advanced sports scenarios, we observed strong performance of unimodal LLMs in zero-shot settings and considerable room for improvement in few-shot learning.
%
Nevertheless, LLMs in general struggle with multi-hop QA tasks. 
This performance disparity emphasizes the potential and the challenges inherent in achieving effective multimodal sports understanding, which integrates both text and vision-based analysis. 
%
%The performance of LLMs in zero-shot advanced scenario analysis indicates that current foundation models possess the potential for addressing complex video sports questions, the results also highlight the substantial challenge for VLMs in video sports understanding compared to LLMs. 
%These findings emphasize the need for LLMs to develop a deeper and more detailed understanding of sports, and for VLMs to gain an in-depth comprehension of specific sports actions.

%\section{Future Work}\vspace{-0.1in}
%Most sports are visually intensive, which underscores the critical role of VLMs in the future development of automated refereeing systems. Although LLMs still have significant room for improvement in few-shot learning for the Advanced Scenario Analysis task, their performance in single-hop indicates that current LLMs already possess a foundational capability for addressing complex video-based sports questions. 

Moving forward, we plan to explore multimodal fine-tuning methods that are tailored for VLMs on the sports understanding domain, by further leveraging the sports dataset overview in this paper.
Inspired by our error analysis, we aim to prioritize interpretability in model decision-making, with a focus on improving cross-modality alignment. Our goal is to contribute to more reliable and transparent automated systems in sports officiating empowered by VLMs.

%Given the indication from the error analysis, we will focus more on the interpretability and contextual understanding of sports videos, 
% The majority of sports are visually driven, making VLMs central to the future development of automated officiating systems. Due to the absence of relevant datasets, we have not established a video QA task that involves complex sports situation reasoning, such as penalty decisions in specific scenarios. While LLMs still have significant space for improvement under few-shot learning in the Advanced Scenario Analysis task, their performance in zero-shot settings demonstrates that current LLMs already provide a foundational capability for answering complex video-based sports questions.

% Looking ahead, we plan to propose a video QA dataset focused on complex sports situation reasoning. This will aid in the pre-training/fine-tuning and comprehensive evaluation of VLMs in Sports Understanding, contributing to their development and enhancing their accuracy and applicability in real-world sports scenarios. This initiative will bridge the gap in current research and provide a robust framework for assessing the capabilities of VLMs in understanding and interpreting intricate sports contexts visually.

\section*{Limitations}
\vspace{-0.1in}
%Despite our intention to encompass a diverse range of LLMs and VLMs in our analysis, 
Financial and computational resource constraints limited our ability to evaluate certain high-capacity models, such as PLLaVA-34B~\citep{xu2024pllava}. %Incorporating models with larger parameter scales may provide deeper insights into their capabilities in Sports Understanding. 
In future work, we aim to broaden the spectrum of models we assess and incorporate models with larger parameter scales to ensure a more comprehensive analysis.
%encompassing a wider range of LLMs and VLMs 

% Bibliography entries for the entire Anthology, followed by custom entries
%\bibliography{anthology,custom}
% Custom bibliography entries only
\bibliography{custom}

\appendix

\section{Appendix}
\label{sec:appendix}

\begin{table*}[htbp!]
%\begin{small}
%\scalebox{0.9}{
\begin{tabular}{p{15cm}}\hline
\textbf{Question:} Question: How might a tennis player effectively use tactics to gain an advantage in a match?\\
A) By consistently serving to the opponent's weak side
B) By hitting deep balls to limit the opponent's shot options
C) By always aiming for the lines to score direct points
D) By varying the pace and spin of the ball to disrupt the opponent's rhythm
\textbf{Answer:} ABCD \\
\textbf{Model Output:}The correct answers are A, B, and D.

Here's why:

A) Serving to the opponent's weak side can be an effective tactic, as it can make it more difficult for them to return the serve and create opportunities for the server to take control of the point.

B) Hitting deep balls can limit the opponent's shot options and make it harder for them to take control of the point. This can be especially effective if the opponent is not comfortable hitting from the back of the court.

C) Aiming for the lines to score direct points is not a reliable tactic, as it's a high-risk strategy that can result in errors and unforced mistakes.

D) Varying the pace and spin of the ball can disrupt the opponent's rhythm and make it harder for them to anticipate and react to the shot. This can be an effective way to create uncertainty and gain an advantage.

Overall, a tennis player can effectively use tactics to gain an advantage in a match by exploiting their opponent's weaknesses, controlling the pace and direction of the ball, and disrupting their opponent's rhythm and timing.
\\
\textbf{Error Classification:} Here are the reasons why it is classified as Context and Nuances Confused:

1. Misinterpretation of effectiveness: The model labeled aiming for the lines as a "high-risk strategy" that could lead to errors and unforced mistakes, suggesting a misunderstanding of the context in which this tactic can be effective. While it is true that aiming for the lines is riskier, it is also a common and effective tactic used in tennis to win points outright.

2. Overemphasis on negative: The model focused solely on the negative aspects of aiming for the lines, failing to recognize or acknowledge that this tactic, when used judiciously, can be a significant advantage in a match. This shows a nuanced misunderstanding of when and how this tactic should be applied.

3. Ignoring professional usage in real world: At higher levels of play, aiming for the lines is a tactic frequently used to apply pressure and secure points, which is crucial for understanding the full context of this tactic's effectiveness.

By not appreciating these nuances, the model failed to see the full strategic value of aiming for the lines, leading to the exclusion of this option from their correct answers. This indicates a confusion of context and nuances surrounding the application of tennis tactics.\\ \hline
\textbf{Question:} Main Question: 
 How does the number of players in a basketball team and the number of substitutes impact the overall game strategy, considering that a standard basketball team comprises 5 players and a professional team can have up to 13 players (including substitutes)?\\
 A. The number of players and substitutes has no impact on the game strategy.
 B. Having more substitutes allows a team to maintain an aggressive playing style without worrying about player fatigue.
 C. The number of players in a team determines the number of substitutes.
 D. A team with more players has a disadvantage as it is harder to coordinate.\\
 Sub-Question 1: 
 What is the standard number of players in a basketball team during a game?\\
 A. 10
 B. 13
 C. 5
 D. 7\\
 Sub-Question 2: 
 Considering a professional basketball team, how many substitutes can a team have?\\
 A. 5
 B. 8
 C. 10
 D. 13 \textbf{Answer:} Main Question: B;Sub-Question1: B;Sub-Question2: B \\
\textbf{Model Output:}Main Question:
The correct answer is B. Having more substitutes allows a team to maintain an aggressive playing style without worrying about player fatigue.

Having a larger roster with more substitutes provides a team with several strategic advantages. It allows them to:\\
\end{tabular}
%}
% \caption{More examples of case study on model response errors.}
\label{tab:morecase1}
%\end{small}
\end{table*}

\begin{table*}[htbp!]
%\begin{small}
%\scalebox{0.9}{
\begin{tabular}{p{15cm}}
Maintain a high level of intensity and aggression throughout the game, without worrying about player fatigue.
Make tactical substitutions to counter the opponent's strategy or to take advantage of mismatches.
Provide rest for key players, allowing them to recover and perform at a higher level.
Experiment with different lineups and rotations to find the most effective combinations.

Sub-Question 1:
The correct answer is C. 5.

In a standard basketball game, each team has 5 players on the court at a time.

Sub-Question 2:
The correct answer is D. 13.

In professional basketball, a team can have up to 13 players on its active roster, including substitutes. This allows teams to have a deep bench and make strategic substitutions throughout the game.
\\
\textbf{Error Classification:} For Sub-Question 1, the model incorrectly recalled the number of people in a professional basketball team can have. The correct number is 13, but the model answered 5, which is the number of starting players.
For Sub-Question 2, the model incorrectly recalled the number of substitutes a professional basketball team can have. The correct number is 8, but the model answered 13, which is actually the total number of players on a professional team, including the starting five.
Therefore, the type of error should be Inaccurate Recall.\\ \hline
\textbf{Question: } Question: How can a player execute a successful attack hit in volleyball?\\
A) By hitting the ball while it’s high above the net.
B) By hitting the ball with force and accuracy towards the opponent's court.
C) By targeting the blockers’ hands and hitting the ball out of bounds.
D) By hitting the ball over the net, aiming for the opponent's weak spots.\\
\textbf{Model Output: } A successful attack hit in volleyball involves a combination of power, accuracy, and strategic placement. Here’s a breakdown of the options:

A) Hitting the ball while it’s high above the net can be part of a successful attack, as it allows the hitter to have a better angle and more options for placement. However, this alone does not guarantee success.

B) Hitting the ball with force and accuracy towards the opponent's court is a key component of a successful attack. Force can make it difficult for the opponent to defend, and accuracy ensures the ball lands in bounds and in a challenging spot for the defense.

C) Targeting the blockers’ hands and hitting the ball out of bounds can be a strategic move known as a "tool" or "wipe." This can be effective if done correctly, as it results in a point for the attacking team.

D) Hitting the ball over the net and aiming for the opponent's weak spots is another crucial aspect of a successful attack. Identifying and targeting areas where the opponent is vulnerable can increase the chances of scoring.

Given these considerations, the most comprehensive answer is:

B) By hitting the ball with force and accuracy towards the opponent's court.
D) By hitting the ball over the net, aiming for the opponent's weak spots.

Both B and D together encompass the essential elements of a successful attack hit in volleyball.\\
\textbf{Error Classification: } The model focused primarily on options B and D, emphasizing force, accuracy, and targeting weak spots. While these are indeed components of a successful attack, the model did not fully acknowledge the importance of the other options (A and C) which are also part of the correct answer. This shows a misunderstanding of the full context of what constitutes a successful attack hit in volleyball. Therefore, this error is classified as Context and Nuances Confused.\\ \hline
\end{tabular}
%}
\caption{More examples of case study on model response errors.}
\label{tab:morecase}
%\end{small}
\end{table*}

\end{document}